\documentclass[11pt,a4paper]{article}
\usepackage[hyperref]{acl2021}
\usepackage{times}
\usepackage{latexsym}

\usepackage[ruled, linesnumbered]{algorithm2e}
\usepackage{graphicx}
\usepackage{amssymb}
\usepackage{amsmath}
\usepackage[ruled, linesnumbered]{algorithm2e}
\usepackage{booktabs}
\usepackage{amsfonts}
\usepackage{multirow}
\usepackage{makecell}
\usepackage{subfigure}
\usepackage{color}
\usepackage{bm}

\makeatletter
\renewcommand{\maketag@@@}[1]{\hbox{\m@th\normalsize\normalfont#1}}%
\makeatother

\usepackage{microtype}

\aclfinalcopy 

\title{CasEE: A Joint Learning Framework with Cascade Decoding for \\ Overlapping Event Extraction}

\author{
Jiawei Sheng$^{1,2}$, Shu Guo$^{3}$, Bowen Yu$^{1,2}$, Qian Li$^{4}$, Yiming Hei$^{5}$,\\
{\bf Lihong Wang}$^{3}$, {\bf Tingwen Liu}$^{1,2}$ and {\bf Hongbo Xu}$^{1,2}$ \\  
$^{1}$Institute of Information Engineering, Chinese Academy of Sciences, Beijing, China\\
$^{2}$School of Cyber Security, University of Chinese Academy of Sciences, Beijing, China\\
$^{3}$National\! Computer\! Network\! Emergency\! Response\! Technical\! Team/Coordination\! Center\! of\! China \\
$^{4}$School of Computer Science and Engineering, Beihang University, Beijing, China \\
$^{5}$School of Cyber Science and Technology, Beihang University, Beijing, China\\
{\tt shengjiawei@iie.ac.cn, guoshu@cert.org.cn, wlh@isc.org.cn}
}

\date{}

\begin{document}
\maketitle

\begin{abstract}
Event extraction (EE) is a crucial information extraction task that aims to extract event information in texts. Most existing methods assume that events appear in sentences without overlaps, which are not applicable to the complicated overlapping event extraction. This work systematically studies the realistic event overlapping problem, where a word may serve as triggers with several types or arguments with different roles. To tackle the above problem, we propose a novel joint learning framework with cascade decoding for overlapping event extraction, termed as CasEE. Particularly, CasEE sequentially performs type detection, trigger extraction and argument extraction, where the overlapped targets are extracted separately conditioned on the specific former prediction. All the subtasks are jointly learned in a framework to capture dependencies among the subtasks. The evaluation on a public event extraction benchmark FewFC demonstrates that CasEE\footnote{The source code is available at \url{https://github.com/JiaweiSheng/CasEE}.} achieves significant improvements on overlapping event extraction over previous competitive methods.
\end{abstract}

\section{Introduction} \label{Sec:Intro}

\begin{figure}[!t]
    \centering
    \includegraphics[width=0.48\textwidth]{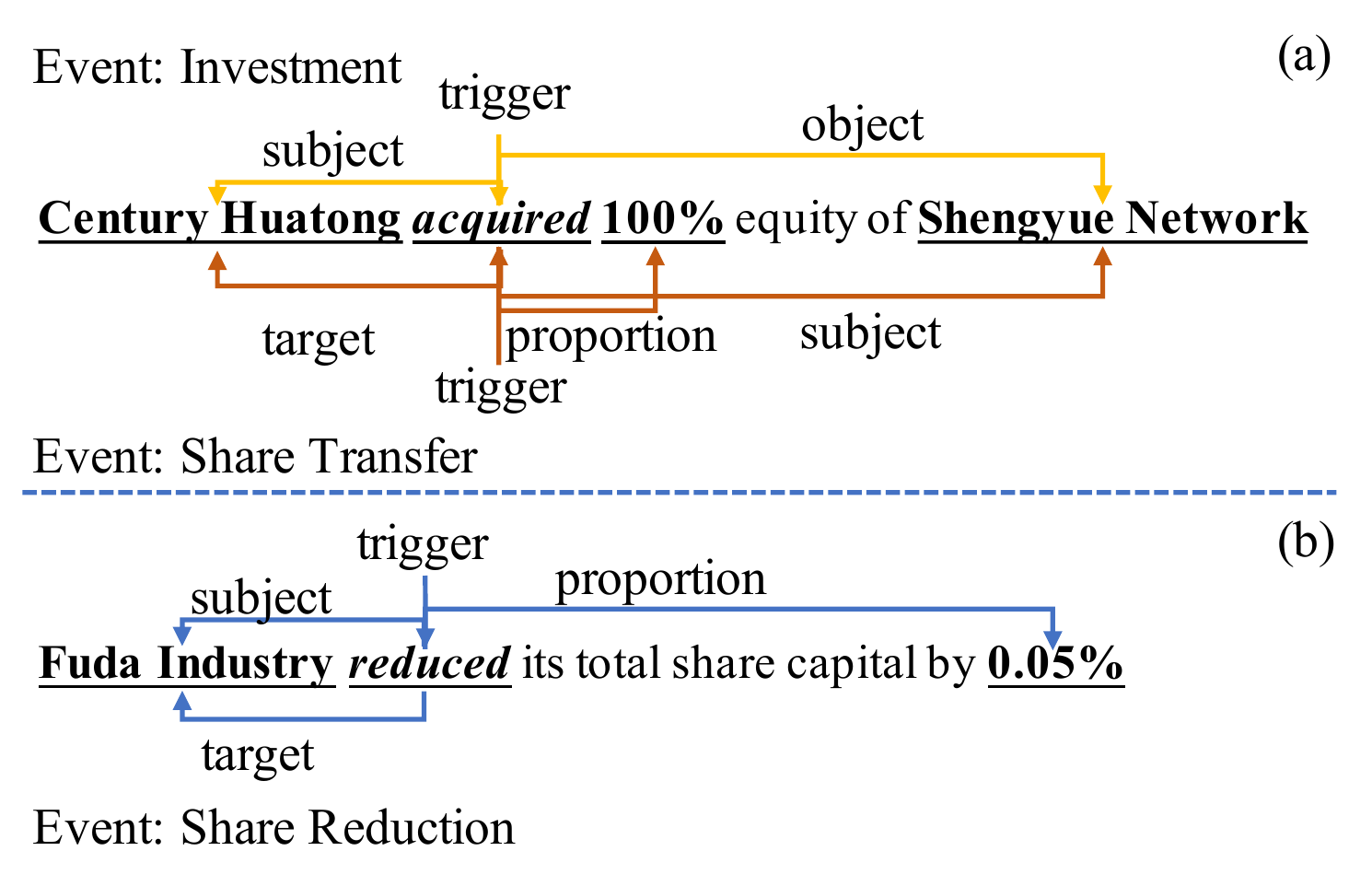}
    \vspace{-6mm}
    \caption{Examples of event overlapping problem: (a) Events with overlapped triggers and arguments; (b) An event with an overlapped argument in several roles.}
    \label{fig:Examples}
\end{figure}

Event Extraction (EE) is an important yet challenging task in natural language understanding. Given a sentence, an event extraction system ought to identify event types, triggers and arguments appearing in the sentence.
As an example, Figure~\ref{fig:Examples}(b) presents an event mention of type {\small \texttt{Share Reduction}}, triggered by ``reduced''. There are several arguments, such as ``Fuda Industry'' playing the {\small \texttt{subject}} role in the event. 

However, events often appear in sentences complicatedly, where the triggers and arguments may have overlaps in a sentence. This paper focuses on a challenging and realistic problem in EE: \textit{overlapping event extraction}. Generally, we categorize all the overlapping cases into three patterns: 
1) A word may serve as triggers with different event types across several events. Figure~\ref{fig:Examples}(a) shows the token ``acquired'' triggers an {\small \texttt{Investment}} event and a {\small \texttt{Share Transfer}} event at the same time. 
2) A word may serve as arguments with different roles across several events. Figure~\ref{fig:Examples}(a) shows ``Shengyue Network'' plays an {\small \texttt{object}} role in the {\small \texttt{Investment}} event and a {\small \texttt{subject}} role in the {\small \texttt{Share Transfer}} event. 
3) A word may serve as arguments playing different roles in one event. Figure~\ref{fig:Examples}(b) shows that ``Fuda Industry'' plays a {\small \texttt{subject}} role and a {\small \texttt{target}} role in an event. For simplicity, we call pattern 1) as \textit{overlapped trigger problem}, and both pattern 2) and 3) as \textit{overlapped argument problem} in the following sections. There are about 13.5\% / 21.7\% sentences having overlapped trigger/argument problems in the Chinese financial event extraction dataset, FewFC~\cite{Yang2021:EAE}. 

Most existing EE studies assume that events appear in sentences without overlaps, which are not applicable to the complicated overlapping scenarios. Typically, current EE studies can be roughly categorized into two groups: 1) Traditional joint methods~\cite{Nguyen16:JRNN,Liu18:JMEE,Nguyen19:Joint3EE}, which simultaneously extract triggers and arguments by a unified decoder labeling the sentence only once. However, they fail in extracting overlapped targets due to label conflicts, where a token may have several typed labels but only one label can be assigned. 2) Pipeline methods~\cite{Chen15:DMCNN,Yang19:PLMEE,Du20:EEQA}, which sequentially extract triggers and arguments in separate stages. \citet{Yang19:PLMEE} attempts to tackle the overlapped argument problem in the pipeline manner, but overlooks the overlapped trigger problem. Nevertheless, the pipeline methods neglect the feature-level dependencies between the trigger and arguments, and suffer from error propagation. In our knowledge, existing researches in EE neglect overlapping problems or only focus on one overlapping problem. Few researches simultaneously solve all the three mentioned overlapping patterns.

To address the above issues, we propose CasEE, a joint learning framework with \underline{Cas}cade decoding for overlapping \underline{E}vent \underline{E}xtraction. Specifically, CasEE realizes event extraction with a shared textual encoder and three decoders for type detection, trigger extraction and argument extraction. To extract overlapped targets across events, CasEE sequentially decodes the three subtasks, conducting trigger extraction and argument extraction according to the former predictions. Such a cascade decoding strategy extracts event elements according to the different conditions, so that the overlapped targets can be extracted in separate phases. A condition fusion function is designed to explicitly model the dependencies between adjacent subtasks. All the subtask decoders are jointly learned to further build connections among subtasks, which refines the shared textual encoder with feature-level interactions among downstream subtasks. 

The contributions of this paper are three-fold:

(1) We systematically investigate the overlapping problems in EE, and categorize them into three patterns. To the best of our knowledge, this paper is among the first to simultaneously tackle all the three overlapping patterns.  

(2) We propose CasEE, a novel joint learning framework with cascade decoding, to simultaneously solve all the three overlapping patterns. 

(3) We conduct experiments on a public Chinese financial event extraction benchmark, FewFC. Experimental results reveal that CasEE achieves significant improvements on overlapping event extraction over existing competitive methods.

\section{Related Work}

Current EE research can be roughly categorized into two groups: 1) Traditional joint methods~\cite{Li2013:JEE_feature,Nguyen16:JRNN,Nguyen19:Joint3EE,Liu18:JMEE,Sha18:BridgeEE} that perform trigger extraction and argument extraction simultaneously. They solve the task in a sequence labeling manner, and extract triggers and arguments by tagging the sentence only once. However, these methods fail in solving overlapping event extraction since the overlapping tokens would cause label conflicts when forced to have more than one label. 2) Pipeline methods~\cite{Chen15:DMCNN,Yang19:PLMEE,Wadden19:DYGIEpp,Li2020:MQAEE, Du20:EEQA, Liu20:RCEE, Chen2020:EEDC} that perform trigger extraction and argument extraction in separate stages. Though pipeline methods have the potential capacity to solve overlapping EE, they usually lack explicit dependencies between triggers and arguments, and suffer from error propagation. Among the researches, \citet{Yang19:PLMEE} and \citet{Xu20:JMCEE} solve the overlapped argument problem, but overlook the overlapped trigger problem, thus can not discern correct triggers for argument extraction. All the above methods can not simultaneously solve all the overlapping patterns in event extraction.

\begin{figure*}[ht]
    \centering
    \includegraphics[width=\textwidth]{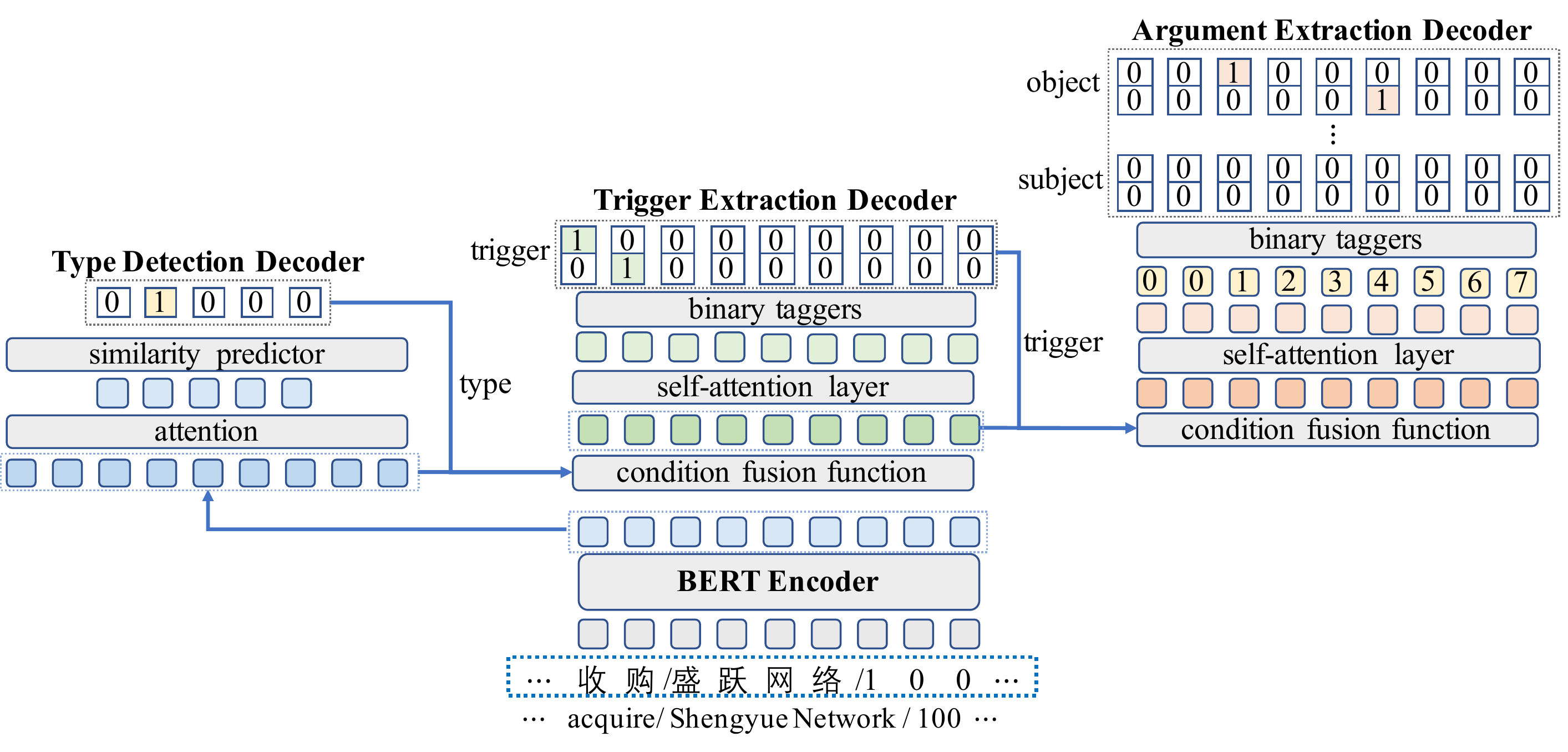}
    \caption{The overview of our proposed approach, CasEE, which contains a shared BERT encoder, a type detection decoder, a trigger extraction decoder and an argument extraction decoder.}
    \label{fig:Model}
\end{figure*}

The overlapping problem has also been explored in other information extraction tasks outside event extraction. \citet{Luo20:BiGraphNestedNER} tackles nested named entity recognization with bipartite flat-graph networks. \citet{Zeng18:seq2seqRTE} tackles overlapped relational triple extraction by applying a sequence-to-sequence paradigm with a copy mechanism. \citet{Wei20:CasRel} and \citet{Yu20:HETER} extract overlapped relational triples with a novel cascade tagging strategy, which inspire us to solve overlapping event extraction in the cascade decoding paradigm. \citet{Wang20:TPLinker} further discusses the propagation error in cascade decoding. All the above researches are proposed for other tasks, which can not be directly transferred for overlapping event extraction due to the complicated event extraction definition.

\section{Our Approach}
Given an input sentence, the goal of EE is to identify triggers with their event types and arguments with their corresponding roles, where triggers and arguments may overlap on some tokens. To tackle this problem, we propose a training objective at the event level. Formally, according to the pre-defined event schema, we have an event type set $\mathcal{C}$ and an argument role set $\mathcal{R}$. The overall goal is to predict all events in gold set $\mathcal{E}_{x}$ of the sentence $x$. We aim to maximize the joint likelihood of training data $\mathcal{D}$: 

\begin{equation}
    \begin{aligned}
    & \prod_{x \in \mathcal{D}} \Big [\prod_{(c,t,a_{r}) \in \mathcal{E}_{x}} p((c,t,a_{r}))|x) \Big ] \\
    = & \prod_{x \in \mathcal{D}} \! \Big [ \! \prod_{c \in \mathcal{C}_{x}} p(c|x) \!\! \prod_{t \in \mathcal{T}_{x,c}} \! p(t|x, c) \!\!\!\!
    \prod_{a_{r} \in \mathcal{A}_{x,c,t}} \!\!\!\! p(a_{r}|x, c, t) \Big]\label{Eq:Overview}
    \end{aligned}
\end{equation}
where $\mathcal{C}_{x}$ denotes the set of types occurring in $x$, $\mathcal{T}_{x,c}$ denotes the trigger set of type $c$, and $\mathcal{A}_{x,c,t}$ denotes the argument set of type $c$ and trigger $t$. Note that each $c$ is a type in $\mathcal{C}$, each $t$ is a trigger word, and each $a_{r} \in \mathcal{A}_{x}$ is an argument word corresponding to its own role $r \in \mathcal{R}$. Eq.~(\ref{Eq:Overview}) exploits the fact of dependencies among the type, trigger and argument. Actually, it motivates us to learn a type detection decoder $p(c|x)$ to detect event types occurring in the sentence, a trigger extraction decoder $p(t|x, c)$ to extract triggers of type $c$, and an argument extraction decoder $p(a_{r}|x, c, t)$ to extract role-specific arguments with type $c$ and trigger $t$.  

Such a task decomposition solves all the event overlapping patterns claimed in the Introduction. Specifically, we first detect event types occurring in the sentence. When extracting triggers, we only predict the triggers with a specific type, thus the triggers overlapped across several events will be predicted in separate phases. Similarly, when extracting arguments, we predict the arguments with a specific type and trigger, thus the arguments overlapped across several events will also be predicted in separate phases. Since we adopt role-specific taggers in argument extraction, the overlapped arguments having several roles in an event can be predicted separately with specific taggers. All the predictions in type detection, trigger extraction and argument extraction form the final prediction. 

Figure~\ref{fig:Model} demonstrates the details of CasEE. CasEE adopts a shared BERT encoder to capture textual features, and three decoders for type detection, trigger extraction and argument extraction. Since all subtasks are jointly learned in contrast to previous pipeline methods~\cite{Yang19:PLMEE, Li2020:MQAEE}, CasEE could capture feature-level dependencies among subtasks. For prediction, CasEE sequentially predicts event types, triggers and arguments in the cascade decoding process.

\subsection{BERT Encoder}

To capture the feature-level dependencies among subtasks, we share the textual representations of each sentence. As BERT has shown performance improvements across multiple NLP tasks, we adopt BERT~\cite{Devlin19:BERT} as our textual encoder. BERT is a bi-directional language representation model based on transformer architecture~\cite{Vaswani2017:Transformer}, which generates textual representations conditioned on token context and remains rich textual information. Formally, the sentence with $N$ tokens is denoted as $x={w_1, w_2, ..., w_{N}}$. We input the tokens into BERT, and then obtain the hidden states $\mathbf{H} = {\mathbf{h}_1, \mathbf{h}_2, ..., \mathbf{h}_{N}}$ as the token representations for the following downstream subtasks. 

\subsection{Type Detection Decoder} \label{Sec:TypeDetector}
Since we tackle the overlapped trigger problem by extracting triggers conditioned on the type predictions, we devise a type detection decoder to predict event types. Inspired by event detection without triggers~\cite{Liu19:TriggerFreeED}, we adopt attention mechanism to detect event types, capturing the most relative context for each possible type. Specifically, we randomly initialize embedding matrix $\mathbf{C}\in \mathbb{R}^{|\mathcal{C}|\times d}$ as the type embeddings. We define a similarity function $\delta$ to measure the relevance between the candidate type $\mathbf{c} \in \mathbf{C}$ and each token representation $\mathbf{h}_{i}$. To fully capture the similarity information in different aspects, we achieve $\delta$ with an expressive learnable function. According to the relevance scores, we obtain the sentence representation $\mathbf{s}_{c}$ adaptive to the type. The details are as follows: 

\begin{equation}
    \begin{aligned}
        \delta (\mathbf{c}, \mathbf{h}_{i})\! &=\! \mathbf{v}^{\intercal} \mathrm{tanh}(\mathbf{W} [\mathbf{c};\mathbf{h}_{i};|\mathbf{c}-\mathbf{h}_{i}|;\mathbf{c} \odot \mathbf{h}_{i}]) \\
        \mathbf{s}_{c} &= \sum_{i=1}^{N} \frac{\mathrm{exp}(\delta (\mathbf{c}, \mathbf{h}_{i}))}{\sum_{j=1}^{N} \mathrm{exp}(\delta (\mathbf{c}, \mathbf{h}_{j}))} \mathbf{h}_{i}
    \end{aligned}
\end{equation}
where $\mathbf{W} \in \mathbb{R}^{4d \times 4d}$ and $\mathbf{v} \in \mathbb{R}^{4d \times 1}$ are learnable parameters, $|\cdot|$ is an absolute value operator, $\odot$ is the element-wise production, and $[\cdot ; \cdot]$ denotes the concatenation of representations.

Finally, we predict event types by measuring the similarity of the adaptive sentence representation $\mathbf{s}_{c}$ and the type embedding $\mathbf{c}$ with the same similarity function $\delta$. Then, the predicted probability of each event type $c$ occurring in the sentence is:

\begin{equation}\label{Eq:TypPre}
    \hat{c} = p(c|x) = \sigma(\delta(\mathbf{c}, \mathbf{s}_{c})) 
\end{equation}
where $\sigma$ denotes sigmoid function. We select the event type with $\hat{c}>\xi_{1}$ as results, where $\xi_{1} \in [0,1]$ is a scalar threshold. All predicted types in sentence $x$ form event type set $\mathcal{C}_{x}$. The decoder learnable parameter $\theta_{td} \triangleq \{\mathbf{W}, \mathbf{v}, \mathbf{C}\}$.

\subsection{Trigger Extraction Decoder}
To discern overlapped triggers with several types, we extract triggers conditioned on a specific type $c\in \mathcal{C}_{x}$. This decoder contains a condition fusion function, a self-attention layer, and a pair of binary taggers for triggers.

To model the conditional dependency between type detection and trigger extraction, we devise a condition fusion function $\phi$ to integrate condition information into textual representation. Specifically, we obtain the conditional token representation $\mathbf{g}_{i}^{c}$ by integrating the type embedding $\mathbf{c}$ into the token representation $\mathbf{h}_{i}$ as:
\begin{equation}\label{Eq:TypedRep}
    \mathbf{g}_{i}^{c} = \phi (\mathbf{c}, \mathbf{h}_{i})
\end{equation}
Actually, $\phi$ can be achieved by concatenation, addition operator or gate mechanism. To fully generate conditional representations in the statistical aspect, we introduce an effective and general mechanism, conditional layer normalization (CLN)~\cite{Su2019:CLN,Yu2021:SOIE}, to achieve $\phi$. CLN is mostly based on the well-known layer normalization~\cite{Ba16:LN}, but can dynamically generate gain $\mathbf{\gamma}$ and bias $\mathbf{\beta}$ based on the condition information. Given a condition embedding $\mathbf{c}$ and a token representation $\mathbf{h}_{i}$, CLN is formulated as: 
\begin{equation}
    \begin{aligned}
        &\mathrm{CLN} (\mathbf{c}, \mathbf{h}_{i}) = \mathbf{\gamma}_{c} \odot (\frac{\mathbf{h}_{i}-\mu}{\sigma}) + \mathbf{\beta}_c, \label{Eq:CLN} \\
        &\mathbf{\gamma}_c = \mathbf{W}_\gamma \mathbf{c} + \mathbf{b}_\gamma,
        \mathbf{\beta}_c = \mathbf{W}_\beta \mathbf{c} + \mathbf{b}_\beta
    \end{aligned}
\end{equation}
where $\mathbf{\mu} \in \mathbb{R}$ and $\mathbf{\sigma} \in \mathbb{R}$ are the mean and standard variance taken across the elements of $\mathbf{h}_{i}$, and $\mathbf{\gamma}_c \in \mathbb{R}^d$ and $\mathbf{\beta}_c \in \mathbb{R}^d$ are the conditional gain and bias, respectively. In this way, the given condition representation is encoded into the gain and bias, and then integrated into contextual representations. 

To further refine representations for trigger extraction, we adopt a self-attention layer over the conditional token representations. Formally, the refined token representations are derived as:

\begin{equation}
    \mathbf{Z}^{c} = \mathrm{SelfAttention}(\mathbf{G}^{c})
\end{equation}
where $\mathbf{G}^{c}$ is the representation matrix composed of $\mathbf{g}_{i}^{c}$. For details of the self-attention layer, please refer to \citet{Vaswani2017:Transformer}.

To predict triggers, we devise a pair of binary taggers. For each token $w_{i}$, we predict whether it corresponds to a start or end position of a trigger as:
\begin{equation}\label{Eq:TriPre}
    \begin{aligned}
        \hat{t}^{sc}_i &= p(t_{s}|w_{i}, c) = \sigma (\mathbf{w}_{t_{s}}^{\intercal} \mathbf{z}_{i}^{c}+b_{t_{s}}) \\
        \hat{t}^{ec}_i &= p(t_{e}|w_{i}, c) =  \sigma (\mathbf{w}_{t_{e}}^{\intercal} \mathbf{z}_{i}^{c}+b_{t_{e}})
    \end{aligned}
\end{equation}
where $\sigma$ denotes sigmoid function, and $\mathbf{z}_{i}^{c}$ denotes the $i$-th token representation in $\mathbf{Z}^{c}$. We select tokens with $\hat{t}^{sc}_i>\xi_{2}$ as the start positions, and those with $\hat{t}^{ec}_i>\xi_{3}$ as end positions, where $\xi_{2},\xi_{3} \in [0,1]$ are scalar thresholds. To obtain the trigger word $t$, we enumerate all the start positions and search the nearest following end position in the sentence, and the tokens between the start and end position form an entire trigger. In this way, the overlapped triggers can be extracted separately according to the type in separate phases. All the predicted trigger $t$ of type $c$ in sentence $s$ forms the set $\mathcal{T}_{c,s}$. The decoder parameter $\theta_{te}$ includes all the parameters in the condition fusion function, the self-attention layer and the trigger taggers.

\subsection{Argument Extraction Decoder}
To tackle the overlapped argument problem, we extract role-specific arguments conditioned on both the specific event type $c \in \mathcal{C}_{s}$ and event trigger $t \in \mathcal{T}_{c,s}$. This decoder also contains a condition fusion function, a self-attention layer, and a group of role-specific binary tagger pairs for arguments. 

We further integrate the trigger information into the typed textual representation $\mathbf{g}_{i}^{c}$ in Eq.~(\ref{Eq:TypedRep}) with function $\phi$ achieved by CLN. Here we take the average of the start and end position token representations of $t$ as the trigger embedding. We also adopt a self-attention layer to derive the refined textual representations $\mathbf{Z}^{ct'}$. To be aware of the trigger position, we adopt the relative position embedding as used in~\citet{Chen15:DMCNN}, which indicates the relative distance from current token to the trigger boundary token. Finally, the token representations $\mathbf{Z}^{ct}$ for argument extraction are derived as:
\begin{equation}
    \mathbf{Z}^{ct} = [\mathbf{Z}^{ct'}; \mathbf{P}]
\end{equation}
where $\mathbf{P} \in \mathbb{R}^{N \times d_{p}}$ is the relative position embeddings, $d_{p}$ is the dimension, and $[\cdot ; \cdot]$ denotes the concatenation of representations.

To predict arguments in roles, we devise a group of role-specific tagger pairs. For each token $w_{i}$, we predict whether it corresponds to a start or end position of an argument of the role $r \in \mathcal{R}$ as:
\begin{equation}\label{Eq:ArgPre}
    \begin{aligned}
        \hat{r}^{sct}_i &= p(a_{r}^{s}|w_{i},c,t) = \mathrm{I}(r, c) \sigma(\mathbf{w}_{r_{s}}^{\intercal} \mathbf{z}_{i}^{ct}+b_{r_{s}}) \\
        \hat{r}^{ect}_i &= p(a_{r}^{e}|w_{i},c,t) = \mathrm{I}(r, c) \sigma(\mathbf{w}_{r_{e}}^{\intercal} \mathbf{z}_{i}^{ct}+b_{r_{e}})
    \end{aligned}
\end{equation}
where $\sigma$ denotes sigmoid function, and $\mathbf{z}_{i}^{ct}$ denotes the $i$-th token representation in $\mathbf{Z}^{ct}$. Since not all roles belonging to the specific type $c$, we adopt an indicator function $\mathrm{I}(r, c)$ to indicate whether the role $r$ belongs to the type $c$ according to the pre-defined event scheme. To make the indicator function derivable, we parameterize $\mathrm{I}(r, c)$ to learn with the model parameters. Specifically, given the type embedding $\mathbf{c} \in \mathcal{C}$, we build the connection between the type and roles as:  
\begin{equation}
    \mathrm{I}(r, c) = \sigma (\mathbf{w}_{r}^{\intercal} \mathbf{c} + b_{r}) 
\end{equation}
where $\sigma$ denotes sigmoid function, $\mathbf{w}_{r},b_{r}$ are parameters associated with the role $r$. For each role $r$, we select tokens with $\hat{r}^{sct}_i>\xi_{4}$ as the start positions, and those with $\hat{r}^{ect}_i>\xi_{5}$ as end positions, where $\xi_{4},\xi_{5} \in [0,1]$ are scalar thresholds. To obtain the argument word $a$ with role $r$, we enumerate all the start positions and search the nearest following end position in the sentence, and the tokens between the start and end position form an entire argument. In this way, the overlapped arguments can be extracted separately according to the different types and triggers with role-specific taggers. All the predicted argument $a_{r}$ with type $c$ and trigger $t$ in sentence $x$ forms the set $\mathcal{A}_{t,c,x}$. The decoder parameter $\theta_{ae}$ includes the type embedding matrix $\mathbf{C}$, and all parameters in the condition fusion function, the self-attention layer and the argument taggers.

\subsection{Model Training}

To train the model, we take log of Eq~(\ref{Eq:Overview}), and the overall objective $\mathcal{J}(\Theta)$ is deployed as:
\begin{equation}
    \begin{aligned}
        \sum_{x\in\mathcal{D}} & \big [ \sum_{c\in\mathcal{C}_{x}} \mathrm{log}\ p_{\theta_{1}}(c|x) + \\ 
        & \sum_{t\in\mathcal{T}_{x,c}} \! \mathrm{log}\ p_{\theta_{2}}(t|x,c)\!\! + \!\!\!\!\!\! \sum_{a_{r}\in\mathcal{A}_{x,c,t}} \!\!\! \mathrm{log}\ p_{\theta_{3}}(a_{r}|x,c,t) \big ]
    \end{aligned}
\end{equation}
where $\Theta \triangleq \{\theta_{1},\theta_{2},\theta_{3}\}$; $p_{\theta_{1}}(c|x)$, $p_{\theta_{2}}(t|x,c)$, and $p_{\theta_{3}}(a_{r}|x,c,t)$ for the subtasks are defined as:
\begin{equation}
    \small
    \begin{aligned}
        &p_{\theta_{1}}(c|x) = (\hat{c})^{\bar{c}}(1-\hat{c})^{(1-\bar{c})}\\
        &p_{\theta_{2}}(t|x,c) = \prod_{z\in\{s,e\}} \prod_{i=1}^{N} (\hat{t}^{zc}_{i})^{\bar{t}^{zc}_{i}}(1-\hat{t}^{zc}_{i})^{(1-\bar{t}^{zc}_{i})}\\
        &p_{\theta_{3}}(a_{r}|x,c,t) \! = \!\! \prod_{r\in \mathcal{R}} \! \prod_{z\in\{s,e\}} \! \prod_{i=1}^{N} (\hat{r}^{zct}_{i})^{\bar{r}^{zct}_{i}}\!(1- \hat{r}^{zct}_{i})^{(1-\bar{r}^{zct}_{i})}
    \end{aligned}
\end{equation}
where $\hat{c}$, $\hat{t}^{sc}_{i}$, $\hat{t}^{ec}_{i}$, $\hat{r}^{sct}_{i}, \hat{r}^{ect}_{i}$ are the predicted probabilities in Eq~(\ref{Eq:TypPre}), Eq~(\ref{Eq:TriPre}), Eq~(\ref{Eq:ArgPre}), and $\bar{c}$, $\bar{t}^{sc}_{i}$,  $\bar{t}^{ec}_{i}$, $\bar{r}^{sct}_{i}$, $\bar{r}^{ect}_{i}$ are the true $0/1$ labels of the training data, respectively. $\theta_{1} \triangleq \{\theta_{bert},\theta_{td}\}$, $\theta_{2} \triangleq \{\theta_{bert},\theta_{te}\}$, $\theta_{3} \triangleq \{\theta_{bert},\theta_{ae}\}$, where $\theta_{bert},\theta_{td}, \theta_{te}, \theta_{ae}$ denote parameters in BERT, type detection, trigger extraction and argument extraction, respectively. We train the model by maximizing $\mathcal{J}(\Theta)$ through Adam stochastic gradient descent~\cite{Kingma2014:Adam} over the shuffled mini-batches.

\section{Experiments}
In this section, we conduct experiments to evaluate the performance of CasEE. 

\subsection{Dataset and Evaluation Metric}

We conduct experiments\footnote{Though ACE 2005 dataset is usually used to evaluate traditional EE models, we observe that it contains a low proportion of sentences with overlapped argument problem (nearly $10\%$ reported in~\citet{Yang19:PLMEE}), and doesn't exist sentences with overlapped trigger problem.} on a Chinese financial event extraction benchmark FewFC~\cite{Yang2021:EAE}. We split data with 8:1:1 for training/validation/testing. Table~\ref{tab:Datasets} shows more details.

\begin{table}[t]
	\centering\footnotesize\setlength{\tabcolsep}{3pt}
	\begin{tabular*}{0.48 \textwidth}{@{\extracolsep{\fill}}@{}l|rrrr@{}}
		\toprule
		      & \#Overlap & \#Normal & \#Sentence  & \#Event\\
		\midrule
		Training    & 1,560     & 5,625  & 7,185    & 10,277\\
	    Validation  & 205       & 694    & 899      & 1,281\\
        Testing     & 210       & 688    & 898      & 1,332\\
        \midrule
	    All         & 1,975     & 7,007  & 8,982   & 12,890\\
        \bottomrule
	\end{tabular*}
	\caption{Statistics of the dataset. Each column denotes the number of the sentences with overlapped elements, the sentences without overlapped elements, all the sentences and all the events.}
	\label{tab:Datasets}
\end{table}

For evaluation, we follow the traditional evaluation metrics~\cite{Chen15:DMCNN,Du20:EEQA}: 1) Trigger Identification (TI): A trigger is correctly identified if the predicted trigger span matches with a golden span; 2) Trigger Classification (TC): A trigger is correctly classified if it is correctly identified and assigned to the correct type; 3) Argument Identification (AI): An argument is correctly identified if its event type is correctly recognized and the predicted argument span matches with a golden span; 4) Argument Classification (AC): an argument is correctly classified if it is correctly identified and the predicted role matches with a golden role. We report Precision (P), Recall (R) and F measure (F1) for each of the four metrics.

\subsection{Comparision Methods}
Though various models have recently been developed for EE, few researches are investigated to solve overlapping event extraction. We attempt to develop the following baselines based on current solutions. For the realistic consideration, no candidate entities are previously known for EE. 

\paragraph{Joint sequence labeling methods.} This kind of method formulates event extraction into a sequence labeling task. {\bf BERT-softmax}~\cite{Devlin19:BERT} adopts BERT to learn textual representations and uses hidden states for classifying event triggers and arguments. {\bf BERT-CRF} adopts conditional random field (CRF) to capture label dependencies, which is adopted in~\cite{DuC20:EventBERTCRF} for document-level event extraction. {\bf BERT-CRF-joint} borrows idea from joint extraction of entity and relation~\cite{Zheng17:JointRTE}, which adopts joint labels of the type and role as {\small \texttt{B/I/O-type-role}}. All the above methods can not solve the overlapping problem due to label conflicts.

\paragraph{Pipelined event extraction methods.} This kind of method solves event extraction with a pipeline manner. {\bf PLMEE}~\cite{Yang19:PLMEE} solves overlapped argument problem by extracting role-specific arguments according to the trigger. Motivated by current Machine Reading Comprehension (MRC) based EE studies~\cite{Li2020:MQAEE, Du20:EEQA, Liu20:RCEE, Chen2020:EEDC}, we train multiple MRC BERTs for overlapping event extraction. We extend MQAEE~\cite{Li2020:MQAEE} for multi-span extraction and re-assemble the following methods\footnote{For more details, please refer to the Appendix B.} to consider conditions in EE: 1) The method first predicts types, and then predicts overlapped triggers/arguments according to the type, termed as \textbf{MQAEE-1}. 2) The method first predicts overlapped triggers with types, and then predicts overlapped arguments according to the typed triggers, termed as \textbf{MQAEE-2}. 3) The method sequentially predicts types, predicts overlapped triggers according to the type, and predicts overlapped arguments according to the type and trigger, termed as \textbf{MQAEE-3}. All the above pipeline methods could solve (or partly solve) overlapping event extraction. 

\begin{table*}[t]
    \centering\footnotesize\setlength{\tabcolsep}{5pt}
    \begin{tabular*}{1 \textwidth}{@{\extracolsep{\fill}}@{}lcccccccccccc@{}}
    \toprule
    \multicolumn{1}{c}{} & \multicolumn{3}{c}{TI(\%)} & \multicolumn{3}{c}{TC(\%)} & \multicolumn{3}{c}{AI(\%)} & \multicolumn{3}{c}{AC(\%)} \\
    \cmidrule{2-4}
    \cmidrule{5-7}
    \cmidrule{8-10}
    \cmidrule{11-13}
    \multicolumn{1}{c}{} & P & R & F1 & P & R & F1 & P & R & F1 & P & R & F1 \\
    \midrule
    BERT-softmax & 89.8  & 79.0  & 84.0  & 80.2 & 61.8  & 69.8  & 74.6  & 62.8  & 68.2 & 72.5 & 60.2 & 65.8 \\ 
    BERT-CRF & \bf 90.8  & 80.8  & 85.5  & \bf 81.7 & 63.6  & 71.5  & 75.1 & 64.3 & 69.3 & 72.9 & 61.8 & 66.9 \\ 
    BERT-CRF-joint & 89.5  & 79.8  & 84.4 & 80.7 & 63.0  & 70.8  & \bf 76.1  & 63.5 & 69.2 & \bf 74.2 & 61.2 & 67.1 \\
    \midrule
    PLMEE & 83.7 & 85.8 & 84.7 & 75.6 & 74.5 & 75.1  & 74.3 & 67.3 & 70.6 & 72.5 & 65.5 & 68.8 \\ 
    MQAEE-1 & 90.1 & 85.5 & 87.7 & 77.3 & 76.0  & 76.6  & 62.9  & 71.5 & 66.9 & 51.7 & 70.4 & 59.6 \\ 
    MQAEE-2 & 89.1 & 85.5 & 87.4  & 79.7 & 76.1  & 77.8  & 70.3  & 68.3  & 69.3 & 68.2 & 66.5 & 67.3 \\ 
    MQAEE-3 & 88.3 & 86.1  & 87.2  & 75.8 & 76.5  & 76.2 & 69.0 & 67.9 & 68.5 & 67.2 & 65.9 & 66.5 \\ 
    \midrule
    CasEE & 89.4 & \bf 87.7 & \bf 88.6 & 77.9 & \bf 78.5 & \bf 78.2 & 72.8 & \bf 73.1 & \bf 72.9 & 71.3 & \bf 71.5 & \bf 71.4    \\ 
    \bottomrule
    \end{tabular*}
    \caption{Results of event extraction on FewFC dataset, where TI, TC, AI, AC denote trigger identification, trigger classification, argument identification and argument classification, respectively.}
    \label{tab:MainResults}
\end{table*}

\subsection{Implementation Details}
We adopt source code for PLMEE with its best hyper-parameters reported in the original literature. To achieve other baselines, we implement the code based on the Transformers library~\cite{wolf2020:transformers}. For all the methods, we adopt Chinese BERT-Base model\footnote{\url{https://huggingface.co/bert-base-chinese}} as the textual encoder, which has 12 layers, 768 hidden units and 12 attention heads. We use the same value for the common hyper-parameters among the methods, including the optimizer, learning rate, batch size and epoch. For all the hyper-parameters, we adopt grid search strategy. We train all the methods with an Adam weight decay optimizer. The initial learning rate is tuned in $[1e^{-5}, 5e^{-5}]$ for BERT parameters and $[1e^{-4},3e^{-4}]$ for other parameters. The warming up proportion for learning rate is $10\%$, and the max training epoch is set to 20. The batch size is set to 8. For CasEE, the dimension $d_{p}$ of the relative position embedding is tuned in $\{16,32,64\}$. To avoid overfitting, we apply dropout to BERT hidden states with the rate tuned in $[0,1]$. Besides, the thresholds $\xi_{1},\xi_{2},\xi_{3},\xi_{4},\xi_{5}$ for prediction are tuned in $[0,1]$. We select the best model leading to the highest performance on the validation data. The optimal hyper-parameter settings are tuned by grid search, listed in the Appendix A. 

\subsection{Main Results}

The performance of all methods on the FewFC dataset is shown in~\ref{tab:MainResults}. The table reveals that:

(1) Compared to the joint sequence labeling methods, CasEE achieves better performance on the F1 score. Specifically, CasEE achieves improvements of 4.5\% over BERT-CRF and 4.3\% over BERT-CRF-joint on F1 score of AC, respectively. Besides, CasEE produces higher results on the recall of the evaluation metrics, since the sequence labeling methods have label conflicts that only one label can be predicted for those multi-label tokens. The results demonstrate the effectiveness of CasEE on overlapping event extraction.

(2) Compared to the pipeline methods, our method also outperforms them on the F1 score. The results show that CasEE achieves 3.1\% and 2.6\% improvements on F1 score of TC and AC over PLMEE, indicating the importance of solving the overlapped trigger problem in EE. Though the MRC based baselines can extract the overlapped triggers and arguments, CasEE still achieves better performance. Specifically, CasEE improves by a relative margin of 4.1\% against the strong baseline MQAEE-2. The reason may be that CasEE jointly learns textual representations for subtasks, building helpful interactions and connections among the subtasks. The results demonstrate the superiority of CasEE over the above pipeline baselines.    

\subsection{Analysis on Overlap/Normal Data}

\begin{table}[t]
    \centering\footnotesize\setlength{\tabcolsep}{5pt}
    \begin{tabular*}{0.48 \textwidth}{@{\extracolsep{\fill}}@{}lc|cccc@{}}
    \toprule
    Variants && TI (\%) & TC (\%) & AI (\%) & AC (\%) \\
    \midrule
    BERT-softmax && 76.5 & 49.0 & 56.1 & 53.5\\
    BERT-CRF && 77.9 & 52.4 & 61.0 & 58.4 \\
    BERT-CRF-joint && 77.8 & 52.0 & 58.8 & 56.8 \\
    \midrule
    PLMEE && 80.7 & 66.6 & 63.2 & 61.4 \\
    MQAEE-1 && 87.0 & 73.4 & 69.4  & 62.3 \\
    MQAEE-2 && 83.6 & 70.4 & 62.1 & 60.1 \\
    MQAEE-3 && 87.5 & 73.7 & 64.3  & 62.2 \\
    \midrule
    Ours &&  \bf 89.0 & \bf 74.9 & \bf 71.5 & \bf 70.3 \\ 
    \bottomrule
    \end{tabular*}
    \caption{Results of overlap sentences in testing. F1 scores are reported for each evaluation metric.}
    \label{tab:overlapping}
  \end{table}

\begin{table}[t]
    \centering\footnotesize\setlength{\tabcolsep}{5pt}
    \begin{tabular*}{0.48 \textwidth}{@{\extracolsep{\fill}}@{}lc|cccc@{}}
    \toprule
    Variants && TI (\%) & TC (\%) & AI (\%) & AC (\%) \\
    \midrule
    BERT-softmax && 86.9 & 79.9 & \bf 76.2 & \bf 74.1 \\
    BERT-CRF && 88.4 & 80.8 & 74.9 & 72.8 \\
    BERT-CRF-joint && 86.9 & 79.9 & 76.1 & 74.0 \\
    \midrule
    PLMEE && 86.4 & 79.7 & 75.7 & 74.0 \\
    MQAEE-1 && 88.0 & 78.5 & 65.1  & 57.7 \\
    MQAEE-2 && \bf 89.0 & \bf 82.0 & 74.2 & 72.3 \\
    MQAEE-3 && 87.1 & 77.6 & 71.3  & 69.6 \\
    \midrule
    Ours &&  88.4 & 80.2 & 74.0 & 72.3 \\ 
    \bottomrule
    \end{tabular*}
    \caption{Results of normal sentences in testing. F1 scores are reported for each evaluation metric.}
    \label{tab:non-overlapping}
\end{table}

To further understand the performance in testing, we divide the original test data into two groups: the sentences with overlapped elements and the sentences without overlapped elements. 

\paragraph{Performance on Overlap Sentences.} As shown in table~\ref{tab:overlapping}, our method significantly outperforms previous methods on the overlap sentences. The improvements may come from the property that our method avoids label conflicts compared to the sequence labeling methods, and builds more effective feature-level connections among subtasks compared to the pipeline methods. 

\paragraph{Performance on Normal Sentences.} As shown in table~\ref{tab:non-overlapping}, our method still performs acceptable results on the normal sentences without overlapped event elements. The sequence labeling methods reveal similiar results on trigger extraction but relatively better results on argument extraction, where they avoid potential propagation errors of the cascade decoding. Besides, PLMEE performs similar results on trigger extraction but relatively better results on argument extraction, where the reason may be that it adopts elaborate re-weighting losses for different argument roles as in its original literature. In addition, MQAEE-2 predicts more accurate triggers since it jointly predicts triggers with types, but it unfortunately ignores feature-level connections among the subtasks, making the argument extraction results similar to CasEE. Even so, the vanilla CasEE still conducts acceptable performance on the normal sentences compared to the baselines. We would further tackle the potential propagation errors and improve the performance for the general event extraction in the future work.

\begin{table}[t]
    \centering\footnotesize\setlength{\tabcolsep}{5pt}
    \begin{tabular*}{0.48 \textwidth}{@{\extracolsep{\fill}}@{}lc|ccc@{}}
    \toprule
    Variants && P (\%) & R (\%) & F1 (\%) \\
    \midrule
    MaxP && 88.1 & 89.2 & 88.5  \\
    MeanP && 88.3 & 89.8 & 88.6 \\
    CLS && \bf 88.9 & 88.7 & 88.5  \\
    \midrule
    CasEE &&  87.5 & \bf 91.9 & \bf 89.2 \\ 
    \bottomrule
    \end{tabular*}
    \caption{Results of type detection decoder variants. The evaluation metric is precision, recall and F1 score in macro average of type classification.}
    \label{tab:ablation1}
\end{table}
\begin{table}[t]
    \centering\footnotesize\setlength{\tabcolsep}{5pt}
    \begin{tabular*}{0.48 \textwidth}{@{\extracolsep{\fill}}@{}lc|ccc@{}}
    \toprule
    Variants && P (\%) & R (\%) & F1 (\%) \\
    \midrule
    w/o self-attention && 89.2 & 88.1 & 88.6   \\
    w/o \ condition && 86.5 & 87.6 & 87.0   \\
    repl. concatenation && 89.3 & 87.7 & 88.5   \\
    repl. addition && \bf 90.4 & 88.8 & 89.6  \\
    repl. gate mechanism && 90.2 & 88.2 & 89.1 \\
    \midrule
    CasEE &&  90.1 & \bf 90.2 & \bf 90.1 \\ 
    \bottomrule
    \end{tabular*}
    \caption{Results of trigger extraction decoder variants. The evaluation metric is precision, recall and F1 score on TC metric with oracle results of type detection.}
    \label{tab:ablation2}
  \end{table}
\begin{table}[t]
    \centering\footnotesize\setlength{\tabcolsep}{5pt}
    \begin{tabular*}{0.48 \textwidth}{@{\extracolsep{\fill}}@{}lc|ccc@{}}
    \toprule
    Variants && P (\%) & R (\%) & F1 (\%) \\
    \midrule
    w/o self-attention &&  82.8 & 81.7 & 82.2   \\
    w/o indicator function && 84.1 & 81.4 & 82.7 \\
    w/o position embedding &&  83.2 & 81.5 & 82.3   \\
    \quad \quad w/o \ condition && \bf 84.7 & 78.2 & 81.3   \\
    \quad \quad repl. concatenation && 84.0 & 79.3 & 81.6   \\
    \quad \quad repl. addition && 84.2 & 78.2 & 81.1  \\
    \quad \quad repl. gate mechanism && 84.6 & 80.2 & 82.4 \\
    \midrule
    CasEE &&  84.1 & \bf 83.7 & \bf 83.9 \\ 
    \bottomrule
    \end{tabular*}
    \caption{Results of argument extraction decoder variants. The evaluation metric is precision, recall and F1 score on AC metric with oracle results of type detection and trigger extraction.}
    \label{tab:ablation3}
\end{table}

\subsection{Discussion for Model Variants}
To investigate the effectiveness of each module, we conduct variant experiments for CasEE. 

\paragraph{Detection Module Variants.} Table~\ref{tab:ablation1} shows performance of type detection variants. Specifically, MaxP/MeanP aggregates textual representations by applying max/mean pooling over BERT hidden states; CLS utilizes the hidden state of the special token {\small \texttt{<CLS>}} as the sentence representation. The results show that our method outperforms all the above variants on F1 score, indicating that learning sentence representation adaptive to the event type produces better representation for type detection.

\paragraph{Extraction Module Variants.} Table~\ref{tab:ablation2} and Table~\ref{tab:ablation3} show performance of decoder variants for trigger extraction and argument extraction, respectively. We remove the self-attention layer in the both extraction decoders, and remove the relative position embeddings and the indicator function in the argument extraction decoder. The results demonstrate the effectiveness of each module. 

Furthermore, we conduct experiments to explore the impact of condition fusion function $\phi$. The experiments include: 1) we simply remove condition integrate function; 2) we achieve $\phi$ by concatenating the condition and token representations; 3) we achieve $\phi$ by simply adding the condition embedding to token representations; 4) we achieve $\phi$ by the gate mechanism, which adds the condition embedding to token representations according to a learnable trade-off factor.  The results show that the performance without condition fusion function decline significantly on the F1 score in the two decoders, since the model can not discern different targets to extract in the sentence. Besides, empirical results also show that CLN performs better performance than other fusion functions on F1 scores in the two decoders, indicating that CLN can generate better conditional token representations for downstream subtasks. 

\section{Conclusion}

This paper proposes a joint learning framework with cascade decoding for overlapping event extraction, termed as CasEE. Previous studies usually assume that events appear in sentences without overlaps, which are not applicable to the complicated overlapping scenarios. CasEE sequentially performs type detection, trigger extraction and argument extraction, where the overlapped targets are separately extracted conditioned on former predictions. All subtasks are jointly learned to capture dependencies among subtasks. Experiments on the public dataset demonstrate that our model outperforms previous competitive methods on overlapping event extraction. Our future work may further tackle the potential error propagation problem in the cascade decoding paradigm, and improve the performance for the general event extraction.     

\section*{Acknowledgments}

Corresponding authors: Lihong Wang and Shu Guo. We thank all the anonymous reviewers for their insightful comments and constructive suggestions. This work is supported by the National Natural Science Foundation of China (No.61772151), and the Youth Innovation Promotion Association of CAS (Grant No. 2021153).

\bibliographystyle{acl_natbib}
\bibliography{anthology,acl2021}

\clearpage

\appendix

\section{Hyper-parameter Settings}
Our implementation is based on PyTorch\footnote{\url{https://pytorch.org/}}. We trained our models with a NVIDIA TESLA T4 GPU. For re-implementation, we report our hyper-parameter settings on the dataset in Table~\ref{tab:hyper-parameters}. 
Note that the hyper-parameter settings are tuned on the validation data by grid search with 3 trials.
\begin{table}[h]
  \centering\footnotesize\setlength{\tabcolsep}{5pt}
  \begin{tabular*}{0.48 \textwidth}{@{\extracolsep{\fill}}@{}lr|r@{}}
  \toprule
  Hyper-parameter &&  CasEE \\
  \midrule
  type embedding dimension $d$             && 768   \\
  position embedding dimension $d_{p}$     && 64   \\
  dropout rate of decoders                 && 0.3  \\
  batch size                               && 8  \\
  training epoch                           && 20  \\
  initial learning rate of BERT            && $2e^{-5}$  \\
  learning rate of decoders                && $1e^{-4}$ \\
  threshold $\xi_{1}$                      && 0.5  \\
  threshold $\xi_{2}$                      && 0.5  \\
  threshold $\xi_{3}$                      && 0.5  \\
  threshold $\xi_{4}$                      && 0.5  \\
  threshold $\xi_{5}$                      && 0.5  \\
  \bottomrule
  \end{tabular*}
  \caption{Hyper-parameter settings of CasEE.}
  \label{tab:hyper-parameters}
\end{table}

\section{Details of MRC Based Baselines}

Here we describe the details of the extended MRC baselines. Since the MRC paradigm could place condition information in the questions, we extend it to solve the overlapping event extraction.

\paragraph{MQAEE-1} contains two models: 1)A BERT classifier to detect event types; 2) A MRC BERT to extract triggers and arguments. The question template is like {\small \texttt{<type>}} to predict triggers with type {\small \texttt{type}}, and {\small \texttt{<role>}} to predict arguments with role {\small \texttt{role}}. Though this method neglects associations between the trigger and argument, it tackles overlapped trigger problem and overlapped argument problem since the overlapped targets are extracted separately according to different questions. 
\paragraph{MQAEE-2} contains two models: 1) A MRC BERT to extract all triggers with types. The question template is a single word {\small \texttt{trigger}} to predict all typed triggers. 2) A MRC BERT to extract arguments in different roles. The question template is like {\small \texttt{<type>and<trigger>}} to predict all arguments associated with the type {\small \texttt{type}} and the trigger {\small \texttt{trigger}}. This method tackles overlapped trigger problem with multiple taggers, and tackles overlapped argument problem by extracting argument separately according to both the type and trigger.
\paragraph{MQAEE-3} contains three models: 1)A BERT classifier to detect event types; 2) A MRC BERT to extract triggers with different types. The question template is like {\small \texttt{<type>}} to predict triggers with type {\small \texttt{type}}. 3) A MRC BERT to extract arguments in different roles. The question template is like {\small \texttt{<type>and<trigger>}} to predict all arguments associated with the type {\small \texttt{type}} and the trigger {\small \texttt{trigger}}. This method tackles overlapped trigger problem by extracting triggers according to the type, and tackles overlapped argument problem according to both the type and trigger.

\end{document}